\documentclass[10pt,twocolumn,letterpaper]{article}

\usepackage[pagenumbers]{cvpr} 
\usepackage{graphicx}
\usepackage{amsmath}
\usepackage{amssymb}
\usepackage{booktabs}
\usepackage{multicol,multirow}
\usepackage{makecell}
\usepackage[table]{xcolor}
\usepackage{xurl}
\usepackage{tcolorbox}
\usepackage{xcolor}
\definecolor{mygreen}{HTML}{3cb44b}
\definecolor{skyblue}{HTML}{beffff}
\definecolor{lightgreen}{HTML}{90ee90}
\newcommand{\PredSty}[1]{\textnormal{\ttfamily\color{mygreen!90!black}#1}\unskip}

\usepackage{color, colortbl}

\usepackage{tabularx}
\usepackage[pagebackref,breaklinks,colorlinks]{hyperref}
\usepackage{colortbl}
\definecolor{mygray}{rgb}{0.89, 0.93, 0.85}
\definecolor{whitesmoke}{rgb}{0.96, 0.96, 0.96}
\definecolor{timberwolf}{rgb}{0.86, 0.84, 0.82}

\usepackage[capitalize]{cleveref}
\crefname{section}{Sec.}{Secs.}
\Crefname{section}{Section}{Sections}
\Crefname{table}{Table}{Tables}
\crefname{table}{Tab.}{Tabs.}
\usepackage{cite}
\usepackage{xspace}

\newcommand{\ours}{Xmodel-VLM\xspace}
\newcommand{\vlm}{Xmodel-VLM\xspace}
\newcommand{\llm}{Xmodel-LM\xspace}

\newcommand{\Hmat}{{\bf H}}

\newcommand{\Xmat}[0]{{{\bf X}}}

\begin{document}

\title{\vlm: A Simple Baseline for Multimodal Vision Language Model}

\author{
Xu Wanting
$\quad$Liu Yang
$\quad$He Langping
$\quad$Huang Xucheng
$\quad$Jiang Ling \\
XiaoduoAI \\
\texttt{\{xuwanting,liuyangfoam,helangping\}@xiaoduotech.com}
}

\maketitle

\begin{abstract}
   We introduce Xmodel-VLM, a cutting-edge multimodal vision language model. It is designed for efficient deployment on consumer GPU servers. Our work directly confronts a pivotal industry issue by grappling with the prohibitive service costs that hinder the broad adoption of large-scale multimodal systems. Through rigorous training, we have developed a 1B-scale language model from the ground up, employing the LLaVA paradigm for modal alignment. The result, which we call Xmodel-VLM, is a lightweight yet powerful multimodal vision language model. Extensive testing across numerous classic multimodal benchmarks has revealed that despite its smaller size and faster execution, Xmodel-VLM delivers performance comparable to that of larger models. Our model checkpoints and code are publicly available on GitHub at \url{https://github.com/XiaoduoAILab/XmodelVLM}.

\end{abstract}

\section{Introduction}
\label{sec:intro}

In recent years, the integration of natural language processing (NLP) and computer vision has spurred significant innovation and breakthroughs within the field of multimodal learning. Notably, advanced visual language models (VLMs) such as GPT-4V~\cite{achiam2023gpt} and Gemini~\cite{gemini} leverage the synergy between text and visual data to achieve advanced understanding and interaction with the world. With their powerful capabilities, they have demonstrated excellent performance in a variety of downstream visual language tasks. At the same time, the rapid development of open source large language models (LLMs) has broken the bottleneck of limited access to technical details of proprietary models, and also laid a solid foundation for the progress of visual language models.

The current mainstream open source visual language models often show excellent performance, which usually relies on the language model components with no less than 7B parameters behind them. For example, Openflamingo~\cite{openflamingo} achieves the best results when using the language model of MPT-7B~\cite{mpt-7b}. BLIP-2~\cite{blip2} uses the 2.7B and 6.7B versions of the OPT~\cite{zhang2022opt} series as well as FlanT5XL (3B) and FlanT5XXL (11B)~\cite{Flan-T5}for exploration, and shows the best performance when the language model is configured as FlanT5XXL (11B) in its architecture. LLaVA-1.5~\cite{LLaVA-1.5} works best when using the language model of Vicuna-13B~\cite{vicuna}. However, the escalating complexity and resource intensity of these large visual language models have cast a spotlight on one of the primary obstacles hindering their widespread adoption: the considerable operational costs. 

In this context, research on small-scale visual language models has become increasingly popular. Existing studies have shown that small-scale visual language models can still perform as well as larger-scale models, such as LLaVA-Phi~\cite{llavaphi}, which combines the open source multi-modal model LLaVA-1.5 and the open source small language model Phi-2(2.7B)~\cite{phi2} to improve multi-modal model resource efficiency. Tiny-LLaVA~\cite{tinyllava} demonstrates that with better training recipes and data quality, smaller-scale Large Multimodal Models (LMMs) can achieve comparable performance to larger LMMs. MobileVLM~\cite{mobilevlm} is designed for efficient operation on mobile devices and includes MobileLLaMA(2.7B), CLIP ViT-L/14~\cite{CLIP}, and lightweight downsample projector (LDP). Table~\ref{tab:VLMs} provides a meticulous comparative overview of the architectures, cross-modal design strategies, and training data sources employed in each model.

Despite the encouraging advancements made in the realm of visual language models, the pursuit of a genuinely optimal harmony between performance and efficiency remains an active and ongoing challenge. To this end, we trained a 1B scale language model (Xmodel-LM-1B \cite{wang2024xmodel}) from scratch, and guided by the cross-modal alignment principle advocated by the LLaVA paradigm, we have delved deeply into various facets of model architecture and training, including the selection of image encoders, the design of image-text connectors, and the exploitation of diverse datasets, all with the aim of pushing the boundaries of what is achievable with a smaller-scale model.

\begin{table*}[ht]
  \centering
  \renewcommand{\arraystretch}{1.2}
  \setlength{\tabcolsep}{1.2pt}
  \scalebox{0.78}{
  \begin{tabular}{llp{3cm}p{3.6cm}p{8.7cm}}
    \toprule
    Model & Vision Encoder & Language Model & Cross-modality Design & Multimodal Training Corpora \\
    \midrule
    CLIP  \cite{CLIP} & ViT, ResNet & Transformer & Linear-Projection & WebImageText~\cite{radford2021learning} (400M image-text pairs) \\
    BLIP  \cite{blip} & ViT & MED & Cross-Attention & 
    COCO \cite{lin2014microsoft}, VG \cite{krishna2017visual}, CC3M~\cite{sharma2018conceptual}, CC12M \cite{changpinyo2021conceptual}, LAION~\cite{schuhmann2021laion} \\
    BLIP-2  \cite{blip2} & CLIP/Eva-CLIP ViT & OPT, Flan-T5  & Q-Former & same as BLIP \\
    InstructBLIP  \cite{dai2023instructblip} & ViT-G/14 & Flan-T5, Vicuna & Q-Former w/ FC & 13 held-in out of 26 public datasets  \\
    Openflamingo  \cite{openflamingo} & CLIP ViT-L/14 & MPT/RedPajama & Cross-Attention & LAION, Multimodal C4~\cite{zhu2024multimodal} \\
    LLaVA  \cite{LLaVA} & CLIP ViT-L/14 & Vicuna 7B/13B & Linear-Projection & filtered CC-595K from CC3M, LLaVA-Instruct-158K \\
    LLaVA-1.5  \cite{LLaVA-1.5} & CLIP ViT-L@336 & Vicuna-7B/13B & MLP  & a subset of InstructBLIP (1.2M)  \\
    MiniGPT-4  \cite{minigpt-4} & EVA-CLIP ViT-G/14 & Vicuna-7B & Q-Former & LAION, CC, SBU  \cite{ordonez2011im2text} \\
    QWen-VL   \cite{Qwen-VL} & Openclip ViT-bigG  \cite{ilharco2021openclip} &  Qwen-7B & Cross-Attention & multi-tasking datasets (Captioning, VQA, Grounding, etc.)\\
    ShareGPT4V  \cite{chen2023sharegpt4v}  & CLIP ViT-L/14@336 & Vicuna-7B & MLP &  ShareGPT4V (100K by GPT-4V, 1.2M by its learned
    model) \\
    LLaVA-Phi  \cite{llavaphi}  & CLIP ViT-L/14@336 & Phi-2-2.7B & MLP  & same as LLaVA-1.5 \cite{LLaVA-1.5} \\
    Tiny-LLaVA  \cite{tinyllava}  & CLIP/SigLIP & \parbox[t]{3.6cm}{TinyLlama \\ StableLM-2 \\ Phi-2} & MLP & same as LLaVA-1.5 and ShareGPT4V \\
    MobileVLM  \cite{mobilevlm}  & CLIP ViT-L/14@336 & MobileLLaMA & LDP & same as LLaVA-1.5 \cite{LLaVA-1.5} \\
    Mini-Gemini  \cite{li2024minigemini}  & CLIP/ConvNeXt~\cite{liu2022convnet} & \parbox[t]{3.6cm}{Gemma-2B \\ Vicuna-7B/13B \\ Mixtral-8x7B \\ Hermes-2-Yi-34B} & MLP & Mini-Gemini-Pretrain, Mini-Gemini-Instruction 
    \\
    \midrule
    \rowcolor{mygray} \ours (ours)  & CLIP ViT-L/14@336 & \llm-1B & XDP & same as LLaVA-1.5 \cite{LLaVA-1.5} \\
    \bottomrule
  \end{tabular}
  }
  \caption{Comparison of open-source VLM architectures and their training corpora.}
  \label{tab:VLMs}
\end{table*}

In this paper, we present \vlm, an innovative vision-language assistant driven by a compact language model. Our contributions are outlined as follows:

\begin{enumerate}
\item We delve into the performance and capabilities of smaller Chinese and English language models, painstakingly trained on terabytes of data.
\item We conduct comprehensive ablation studies on the design of \vlm, meticulously evaluating various design choices and their respective impacts.
\end{enumerate}

These contributions not only shed light on the efficacy of compact language models but also offer insights into optimizing the design of vision-language models for enhanced performance and efficiency.

\section{Related Work}
\label{sec:related}
\subsection{Vision Transformer} 

\begin{figure}[ht]
  \centering
   \includegraphics[width=\linewidth]{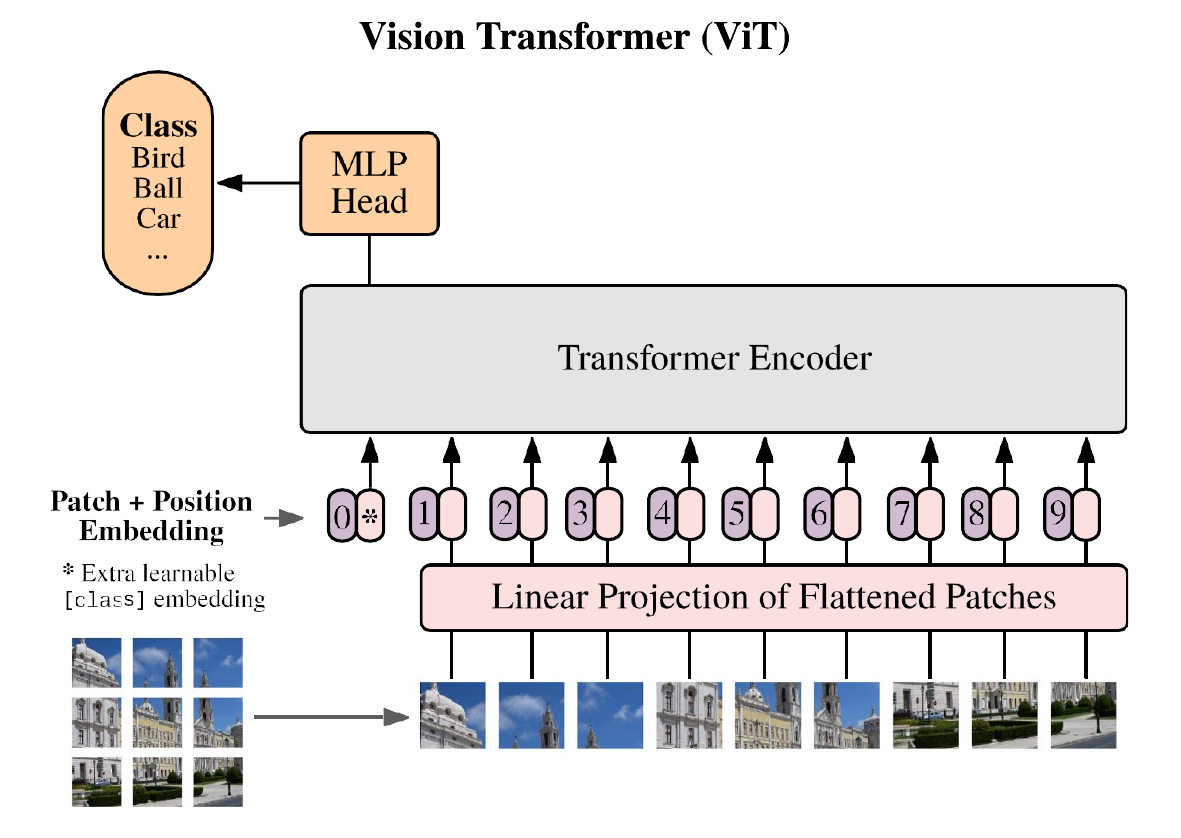}
   \caption{Vision Transformer}
   \label{fig:vision-transformer-arch}
\end{figure}

The Vision Transformer (ViT) represents a pioneering model that adopts a Transformer-like architecture applied to patches extracted from the image. Initially, the image is segmented into fixed-size patches, followed by linear embedding of each patch. Position embeddings are subsequently incorporated, and the resultant sequence of vectors is fed into a standard Transformer architecture, as depicted in Figure~\ref{fig:vision-transformer-arch}. This innovative approach marks a departure from conventional convolutional neural networks (CNNs), offering promising avenues for enhanced image understanding and representation learning.

\subsection{LLaVA} 
LLaVA represents a pioneering approach to constructing cost-effective universal multimodal assistants. It embodies a novel end-to-end trained large-scale multimodal model that ingeniously integrates visual encoders with Vicuna~\cite{vicuna} for comprehensive vision and language understanding. Inspired by the multimodal GPT-4 framework, LLaVA not only excels in conversational interactions but also achieves state-of-the-art accuracy in Science QA tasks.

LLaVA leverages ChatGPT/GPT-4 for multimodal instruction-following data collection, taking advantage of abundant image-pair data. Given an image $\mathbf{X_v}$ and its caption $\mathbf{X_c}$, the model prompts GPT-4 to genenrate a series of questions aimed at instructing the assistant to describe the image content. Consequently, an image-text pair is expanded into its instruction-following version, as illustrated below:

$
\texttt{Human}: \Xmat_{\texttt{q}} ~\Xmat_{\texttt{v}}     \texttt{<STOP>}~
    \texttt{Assistant}: 
    \Xmat_{\texttt{c} } 
     \texttt{<STOP>}
$. 

\begin{table*}[t!]\centering
\begin{minipage}{1.99\columnwidth}\vspace{0mm}    \centering
\begin{tcolorbox} 
    \raggedright
    \small
     \hspace{-6mm}

    $\Xmat_{\texttt{system-message}}$  \PredSty{\texttt{<STOP>}} \\
    $\texttt{Human}: \Xmat_{\texttt{instruct}}^1$  \PredSty{\texttt{<STOP>}}
    $\texttt{Assistant}$: 
    \PredSty{$\Xmat_{\texttt{a} }^1$} 
    \PredSty{\texttt{<STOP>}} \\
    $\texttt{Human}: \Xmat_{\texttt{instruct}}^2$  \PredSty{\texttt{<STOP>}}
    $\texttt{Assistant}$: 
    \PredSty{$\Xmat_{\texttt{a} }^2$} 
    \PredSty{\texttt{<STOP>}} $\cdots$ \\  

\end{tcolorbox}
    
\vspace{-2mm}
\caption{Input sequence used for model training. Only two conversation turns are shown here; in practice, the number of turns varies depending on the instruction-following data. The model is trained to predict assistant responses and determine stopping points, with only {\color{mygreen} green sequences/tokens} used to compute loss in the auto-regressive model.}
    \label{tab:input_sequence}
\vspace{-5mm}
\end{minipage}
\end{table*}

The sequence above illustrates the unified format for the multimodal instruction-following sequence. This transformative process elevates conventional image-text pairing into a comprehensive instructional scenario, facilitating deeper cross-modal understanding and responsiveness in the model.

For each input image $\mathbf{X_v}$, the vision encoder $g$ extracts visual features $\mathbf{Z_v}=g(\mathbf{X_v})$. Subsequently, a trainable projection network converts $\mathbf{Z_v}$ into language embedding tokens $\mathbf{H_v}$, aligning their dimensions with the word embedding space in the language model:

\begin{equation}
\mathbf{H_v}=\mathbf{P}(\mathbf{Z_v}), \quad \text{with} ; \mathbf{Z_v}=g(\mathbf{X_v})
\end{equation}

Thus, a sequence of visual tokens $\mathbf{H_v}$ is obtained. For generating multi-turn conversation data $(\mathbf{X_q}^1,\mathbf{X_a}^1,\cdots,\mathbf{X_q}^T,\mathbf{X_a}^T)$ corresponding to each image $\mathbf{X_v}$, all answers are considered as the assistant’s responses, resulting in a unified structure for the multimodal instruction-following sequence. The LLM then instruction-tuned on prediction tokens using its original auto-regressive training objective.

Specifically, for a sequence of length $L$, the probability of target answers $\mathbf{X_a}$ is computed as follows:

\begin{equation}
p(\mathbf{X_a} | \mathbf{X_v}, \mathbf{X_q}) =
\prod_{i=1}^{L} p_{\theta} ( {\color{mygreen} x_i} | \mathbf{X_v}, \mathbf{X_{q, <i}}, \mathbf{X_{a, <i}})
\label{eq:auto_regressive}
\end{equation}

where $\theta$ represents trainable parameters, and $\mathbf{X_{q, <i}}$ and $\mathbf{X_{a, <i}}$ denote instruction and answer tokens preceding the current prediction token ${\color{mygreen} x_i}$, respectively.

The LLaVA model training involves a two-stage instruction-tuning procedure:

\paragraph{Stage 1: Pre-training for Feature Alignment.}
By utilizing 595K image-text pairs filtered from CC3M, which are reformatted into an instruction-following format, this stage aims to align image features $\Hmat_{\texttt{v}}$ with the pre-trained LLM word embeddings. During this stage, the visual encoder and the LLM weights remain frozen. Training is focused on the trainable parameters within the projector. This is equivalent to create a visual tokenizer compatible with the frozen LLM embeddings.

\paragraph{Stage 2: Fine-tuning End-to-End.}
The visual encoder weights remain frozen while continuing to update both pre-trained projector and LLM weights.

\subsection{MobileVLM}
\begin{figure}[ht]
  \centering
   \includegraphics[width=0.6\linewidth]{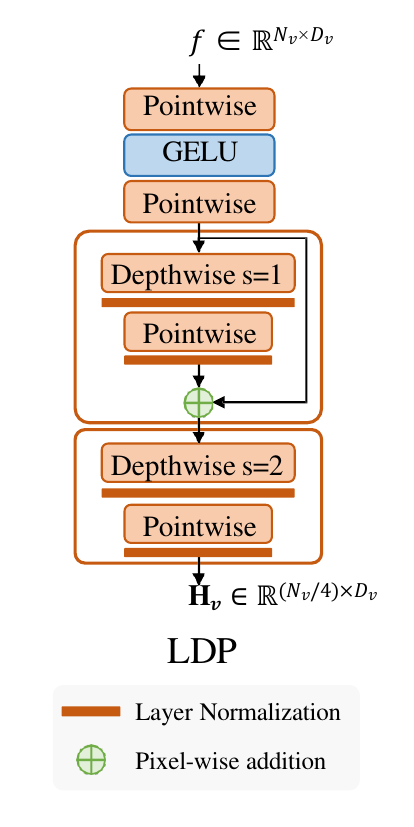}
   \caption{LDP architecture}
   \label{fig:ldp-arch}
\end{figure}

MobileVLM is a multimodal vision language model designed explicitly for deployment on mobile devices. Central to its efficiency is the Lightweight Downsample Projector (LDP), as depicted in Figure~\ref{fig:ldp-arch}. The LDP reduces the number of visual tokens by 75\%, leading to a significant increase in inference speed.

\section{Model architecture}
\label{sec:method}

\subsection{Overall Architectural Design}

\begin{figure}[ht]
  \centering
   \includegraphics[width=\linewidth]{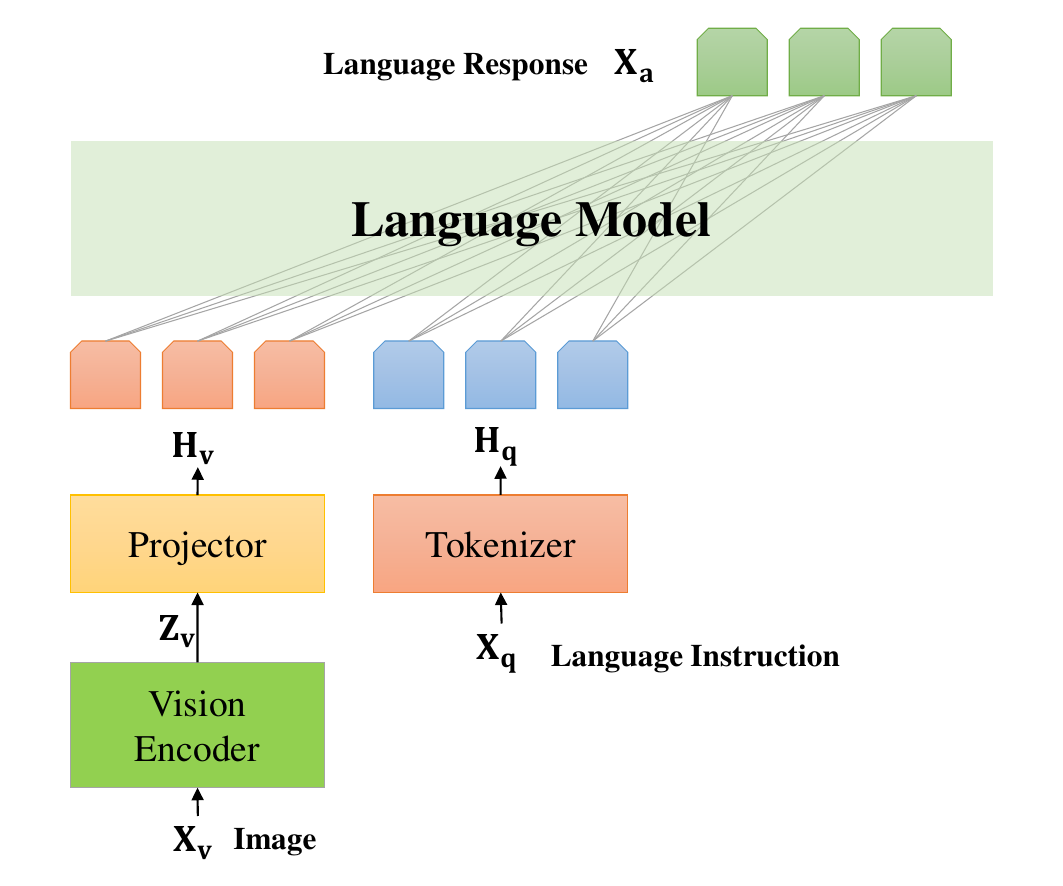}
   \caption{\vlm architecture}
   \label{fig:xmodelvlm-arch}
\end{figure}

The architecture of our network, illustrated in Figure~\ref{fig:xmodelvlm-arch}, closely mirrors that of LLaVA-1.5. It consists of three key components: 1) a vision encoder, 2) a lightweight language model (LLM), and 3) a projector responsible for aligning the visual and textual spaces.

\subsection{Vision Encoder}
Our model utilizes the pre-trained CLIP ViT-L/14 with a resolution of 336×336 as the visual encoder.

\subsection{Language Model}
To reduce operational costs, we trained an lightweight language model \llm 1.1B  from scratch. To ensure seamless integration with existing inference frameworks, our language model is designed to closely resemble LLaMA. The detailed settings of our LLM are provided in Table~\ref{tab: LLM_setting}. 

\begin{table}[ht]
  \centering
  \setlength{\tabcolsep}{2pt}
  \begin{tabular}{@{}lccccc@{}}
    \toprule
    \# of Params & Hidden size & Heads & Layers &    Context Len\\
    \midrule
    1.1B &2048& 32 & 24 &  4096\\
    \bottomrule
  \end{tabular}
  \caption{Detailed settings of our language model.}
  \label{tab: LLM_setting}
\end{table}

To tokenize text data, we employ the unigram algorithm~\cite{unigram}, utilizing the implementation provided by SentencePiece~\cite{sentencepiece}.

In Table~\ref{tab:llm-comparison},  we evaluate our model using standard natural language benchmarks, focusing on language understanding and common sense reasoning. The evaluation is conducted utilizing the Language Model Evaluation Harness tool\cite{eval-harness}. Our experimental results demonstrate that our \llm 1.1B model performs comparably to recent open-source models, including OPT 1.3B, Pythia 1.4B, MobileLLaMA 1.4B and TinyLLaMA 1.1B.

\begin{table*}[ht]
  \centering
  \setlength{\tabcolsep}{3pt}
  \scalebox{0.93}{
  \begin{tabular}{l*{12}{c}}
    \toprule
    & ARC$_{(c/e)}$ & BoolQ & HellaSwag & OpenBookQA &  PIQA & SciQ & TriviaQA & Winogrande  & Avg. \\
    \midrule
    OPT 1.3B \cite{zhang2022opt} & 23.29 / 57.07 & 57.89 & 41.52 & 23.40 & 71.65 &84.20 & 7.43 & 59.51 &47.33 \\
    Pythia 1.4B \cite{biderman2023pythia} & 26.02 / 60.69 & 63.09 &40.43  & 22.40 & 70.73 &86.40& 5.67 & 57.30 & 48.08\\
    MobileLLaMA 1.4B \cite{mobilevlm} & 26.28 / 61.32 & 57.83 &42.87  & 23.60 & 71.33 &87.30 &12.02 & 58.25 & 48.98\\
    TinyLLaMA 1.1B (3T) \cite{zhang2024tinyllama}& 27.82 / 60.31& 57.83 &44.98 & 21.80 & 73.34 &88.90& 11.30 & 59.12& 49.49   \\
    \midrule
    \rowcolor{mygray}\llm 1.1B & 28.16 / 62.29& 61.44& 45.96& 24.00 & 72.03&89.70& 18.46& 60.62 &51.41  \\
    \bottomrule
  \end{tabular}
  }
  \caption{Comparison with 1B scale state-of-the-art language models on prominent language benchmarks}
  \label{tab:llm-comparison}
\end{table*}

\begin{table*}[ht]
\centering
\setlength{\tabcolsep}{3pt}
\scalebox{0.97}{
\begin{tabular}{*{1}{l}*{2}{l}|*{9}{cc}}
\toprule
Method & LLM & Res. & VizWiz & SQA$^\text{I}$ & VQA$^\text{T}$ & POPE & GQA & MMB & MMB$^\text{CN}$ & MM-Vet & MME  \\
\midrule
Openflamingo \cite{openflamingo} & MPT-7B & 336 & -- & -- & 33.6 & -- & -- & 4.6 & -- & -- & -- \\
BLIP-2 \cite{blip2} & Vicuna-13B & 224 & 19.6 & 61.0 & 42.5 & 85.3 & 41.0 & -- & -- & 22.4 & 1293.8 \\
MiniGPT-4 \cite{minigpt-4} & Vicuna-7B & 224 & -- & -- & -- & -- & 32.2 & 23.0 & -- & 22.1 & 581.7\\
InstructBLIP \cite{dai2023instructblip} & Vicuna-7B & 224 & 34.5 & 60.5 & 50.1 & -- & 49.2 & 36 & 23.7 & 26.2 & -- \\
InstructBLIP \cite{dai2023instructblip}& Vicuna-13B & 224 & 33.4 & 63.1 & 50.7 & 78.9 & 49.5 & -- & -- & 25.6 & 1212.8\\
Shikra \cite{chen2023shikra}& Vicuna-13B & 224 & -- & -- & -- & -- & -- & 58.8 & -- & -- & --\\
Qwen-VL \cite{Qwen-VL} & Qwen-7B & 448 & 35.2 & 67.1 & 63.8 & -- & 59.3 & 38.2 & 7.4 & -- & 1487.6\\
MiniGPT-v2 \cite{minigpt-v2} & LLaMA-7B & 448 & -- & -- & -- & -- & 60.3 & 12.2 & -- & -- & --\\
LLaVA-v1.5-13B \cite{LLaVA-1.5} & Vicuna-13B & 336 & 53.6 & 71.6 & 61.3 & 85.9 & 63.3 & 67.7 & 63.6 & 35.4 & 1531.3 \\
MobileVLM 1.7  \cite{gemini} & MobileLLaMA 1.4B & 336 & 26.3 & 54.7 & 41.5 & 84.5 & 56.1 & 53.2 & 16.67 & 21.7 & 1196.2 \\
\midrule
\rowcolor{mygray}
\vlm & \llm 1.1B & 336 & 41.7 & 53.3 & 39.9 & 85.9 & 58.3 & 52.0 & 45.7 & 21.8 & 1250.7 \\
\bottomrule
\end{tabular}
}
\caption{\textbf{Comparison with SOTA methods on 9 VLM benchmarks.} VizWiz \cite{gurari2018vizwiz}; SQA$^\text{I}$: ScienceQA-IMG  \cite{lu2022learn}; VQA$^\text{T}$: TextVQA  \cite{singh2019vqa}; POPE  \cite{li2023evaluating}; GQA  \cite{hudson2019gqa}; MMB: MMBench \cite{liu2024mmbench}; MMB$^\text{CN}$: MMBench-Chinese \cite{liu2024mmbench}; MM-Vet \cite{yu2023mm}; MME  \cite{fu2024mme}; Column \textit{Res.} is the image resolution of vision model. 
}
\label{tab:compare-with-sotas-vlms}
\end{table*}

\subsection{Projector}
\label{Projector}

\begin{figure}[ht]
  \centering
   \includegraphics[width=\linewidth]{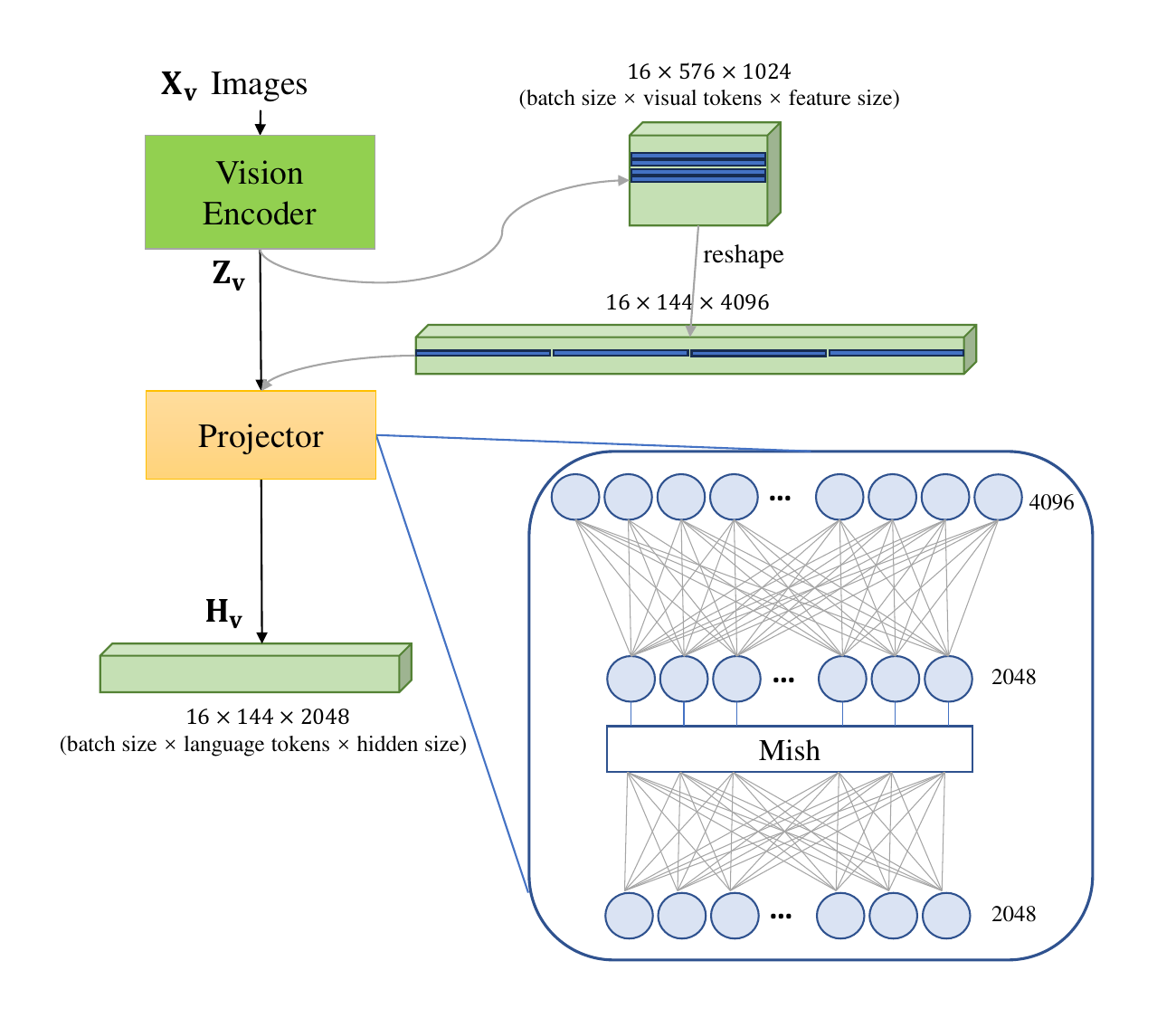}
   \caption{XDP architecture}
   \label{fig:xdp}
\end{figure}

We employ a two-layer MLP to enhance the connection between the vision encoder and LLM, and use the Mish~\cite{misra2019mish} function for activation. Notably, this innovative projector also serves as a downsampling mechanism, effectively reducing the number of visual tokens by 75\%. Our projector architecture, denoted as XDP, exemplifies a paradigm of simplicity and efficacy, as shown in Figure~\ref{fig:xdp}. 

\section{Experiment}
\label{sec:exp}

\subsection{Training}

\begin{figure}[ht]
\centering
\includegraphics[width=\linewidth]{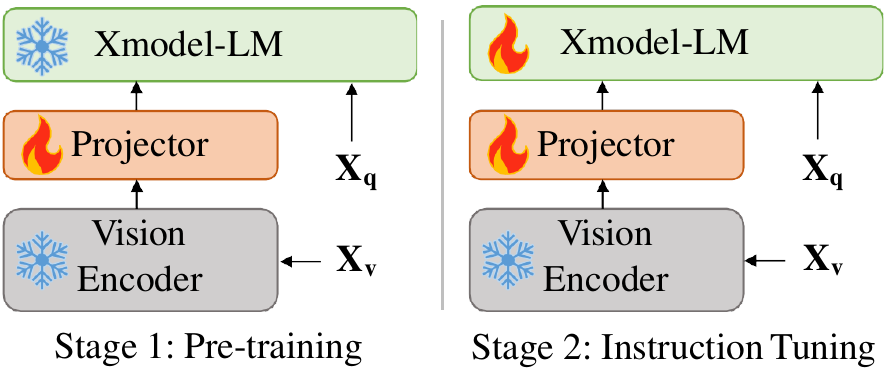}
\caption{Training strategy}
\label{fig:training-strategy}
\end{figure}

Our training process involves two main steps: pre-training and instruction tuning, as illustrated in Figure~\ref{fig:training-strategy}.

Initially, we focus on training efficient projector while keeping the visual encoder and \llm frozen. Subsequently, we conduct comprehensive fine-tuning of both the projector and the language model (LLM) to enhance their visual understanding and language processing capabilities.

For pretraining, we utilize a filtered subset of the CC-595K dataset for one epoch, with an initial learning rate of 1e-3 and a batch size of 64. We then fine-tune the model for one epoch on the LLaVA-Instruct-150K dataset, using a learning rate of 4e-5 and a batch size of 32. We employ a weight decay of 0 and utilize the AdamW optimizer with momentum parameters of 0.9 and 0.999, along with an epsilon value of 1e-8. During fine-tuning, all parameters in LLM are updated without using LoRA.

\subsection{Evaluation of \vlm}

We evaluate the multimodal performance across a variety of datasets: VizWiz \cite{gurari2018vizwiz}, SQA$^\text{I}$ \cite{lu2022learn}, VQA$^\text{T}$ \cite{singh2019vqa}, POPE \cite{li2023evaluating}, GQA \cite{hudson2019gqa}, MMB \cite{liu2024mmbench}, MMB$^\text{CN}$ \cite{liu2024mmbench}, MM-Vet \cite{yu2023mm}, and MME \cite{fu2024mme}. Our analysis, as depicted in Table~\ref{tab:compare-with-sotas-vlms}, illustrates that despite having a smaller parameter count, our proposed \vlm 1.1B demonstrates competitive performance.



We evaluated the inferential latency of our model, comparing it with LLAVA-7B and Mobile-VLM models. Utilizing the SQA$^\text{I}$ dataset on a single NVIDIA GeForce RTX 3090 GPU with 24GB memory, we measured inference speed, excluding preprocessing time. Results in Table~\ref{tab:lantency comparison} show our model's faster inference compared to LLAVA-7B, although MobileVLM-1.7B remains faster. This highlights the advantage of compact models in delivering expedited inference results.

\begin{table}[ht]
  \centering
  \setlength{\tabcolsep}{3pt}
  \scalebox{0.97}{
  \begin{tabular}{@{}lccccc@{}}
  \toprule
Model     & Size(GB) & Samples(token/s) & Total(s) &  &  \\ 
\midrule
LLaVA-7B & 7B       & 19.06               & 1090.32         &  &  \\ 
MoblieVLM-1.7B &   1.7B       & 919.25               & 579.88         &  &  \\ 
\midrule
\rowcolor{mygray}\vlm    & 1.1B        & 415.69               & 1360.25         &  &  \\ 
\midrule
  \end{tabular}
  }
  \caption{Lantency comparison of small VLMs. Smaples reflects a model can predict the number of token per second, and Total denotes the total time of inference on the SQA$^\text{I}$ dataset.}
  \label{tab:lantency comparison}
\end{table}

\section{Ablation Study}




\subsection{Projectors} 
With the language model fixed to \llm 1.1B, vision encoder to clip-vit-large-patch14-336, and token number to 144, we examine different projector architectures and their impact on multimodal performance, as depicted in Table~\ref{tab:quantitive_of_projector_arch}.

\begin{table}[h]
\centering
\setlength{\tabcolsep}{3pt}
\begin{tabular}{*{1}{l}|*{6}{c}}
\toprule
Projector  & GQA & SQA$^\text{I}$ & VQA$^\text{T}$ & POPE  & MME & MMB$^\text{dev}$\\
\midrule
Linear  & 59.2 & 31.6 & 38.7 & 85.7 & 1274.0 & 42.7 \\
MLP & 60.1  & 49.2  & 40.1 & 86.8 & 1294.1 & 49.1 \\
LDP & 57.5  & 52.5  & 37.2 & 85.5 & 1230.5 & 45.3 \\
LDPv2 & 58.8  & 52.8  & 38.4 & 86.2 & 1275.6 & 48.6 \\
\midrule
\rowcolor{mygray}XDP & 58.3  & 53.3  & 39.9 & 85.9 & 1250.7 & 52.0 \\
\bottomrule
\end{tabular}
\caption{Comparison with different projector architectures}
\label{tab:quantitive_of_projector_arch}
\end{table}

\subsection{Token Numbers} 
In order to explore the impact of different visual token numbers on model performance, we keep the language model as \llm 1.1B, vision encoder as clip-vit-large-patch14-336, and projector architecture as XDP, we change the number of tokens and assess multimodal performance, as illustrated in Table~\ref{tab:quantitive_of_token_numbers}. 

\begin{table}[h]
\centering
\setlength{\tabcolsep}{3pt}
\begin{tabular}{*{1}{l}|*{6}{c}}
\toprule
Tokens  & GQA & SQA$^\text{I}$ & VQA$^\text{T}$ & POPE  & MME & MMB$^\text{dev}$\\
\midrule
576  & 60.1 & 53.6 & 40.3 & 86.9 & 1259.4 & 48.3 \\
288 & 59.1  & 42.8  & 39.8 & 86.8 & 1226.2 & 48.5 \\
\rowcolor{mygray}144 & 58.3  & 53.3  & 39.9 & 85.9 & 1250.7 & 52.0 \\
72 & 57.0  & 46.4  & 37.7 & 84.9 & 1234.0 & 45.3 \\
64 & 56.9  & 43.6  & 35.3 & 84.9 & 1234.7 & 43.6 \\
36 & 55.2  & 52.9  & 33.8 & 84.3 & 1214.0 & 47.1 \\
18 & 54.4  & 44.7  & 32.1 & 83.3 & 1247.1 & 40.2 \\
8 & 52.3  & 40.8  & 29.8 & 80.7 & 1120.5 & 41.2 \\
4 & 51.2  & 43.2  & 29.2 & 80.1 & 1040.1 & 36.2 \\
2 & 49.7  & 44.0  & 28.8 & 79.6 & 1006.6 & 33.0 \\
1 & 49.4  & 44.7  & 27.5 & 79.2 & 1070.6 & 32.1 \\
\bottomrule
\end{tabular}
\caption{Comparison with different token numbers}
\label{tab:quantitive_of_token_numbers}
\end{table}

Surprisingly, as the token count decreases, although there is an overall downward trend in various evaluation metrics, the change is remarkably gradual. Especially when only a small number of visual tokens are used, such as 1 and 2, there is no significant difference in the evaluation results. Of course, this may be related to the fact that there is no fine-grained description of the image in the datasets, which is the limitation of the vision encoder in modality alignment.

\subsection{Language Model} 
To investigate the potential of our approach, we compared the impact of different sizes of LLM on performance based on the structure of the vision encoder as clip-vit-large-patch14-336 and projector architecture as XDP with 144 tokens, and the results are shown in Table~\ref{tab:quantitive_of_language_model}. This result shows that larger language models are an effective way to improve the performance of our model, which provides a direction for further improvement of our method in the future.   

\begin{table}[h]
\centering
\setlength{\tabcolsep}{1.5pt}
\scalebox{0.93}{
\begin{tabular}{*{1}{l}|*{6}{c}}
\toprule
LLM  & GQA & SQA$^\text{I}$ & VQA$^\text{T}$ & POPE  & MME & MMB$^\text{dev}$\\
\midrule
LLaMA-3 8B & 60.8  & 68.8  & 51.4 & 86.7 & 1458.7 & 65.0 \\
Phi-3 4B & 59.4  & 71.2  & 51.3 & 86.8 & 1388.7 & 65.8 \\
\midrule
\rowcolor{mygray}\llm 1.1B & 57.4  & 54.4  & 38.9 & 86.1 & 1251.5 & 48.5 \\
\bottomrule
\end{tabular}
}
\caption{Comparison with different Language Models}
\label{tab:quantitive_of_language_model}
\end{table}



\section{Conclusion}
\label{sec:concl}

In summary, we present a high-performance visual language model achieved through careful selection of visual encoders, efficient projector design, and a two-stage training strategy. Extensive experiments on popular VLM benchmarks demonstrate its effectiveness. We anticipate our technology will unlock new possibilities across various applications, including customer service robots.

\clearpage
\clearpage
{\small
\bibliographystyle{ieee_fullname}
\bibliography{Xmodel-VL}
}

\clearpage
\clearpage
\appendix

\end{document}